\documentclass{amia}
\usepackage{graphicx}
\usepackage[labelfont=bf]{caption}
\usepackage{amsfonts, amsmath, amsthm, amssymb, bbm}
\usepackage[superscript,nomove]{cite}
\usepackage{color}
\usepackage{amsmath}
\usepackage{amssymb}
\usepackage{multirow}
\usepackage{hyperref}
\usepackage{setspace}

\begin{document}

\title{DEPLOYR: A technical framework for deploying custom real-time machine learning models into the electronic medical record}

\author{Conor K. Corbin$^{*,1}$, Rob Maclay$^{*,2}$, Aakash Acharya$^{3}$, Sreedevi Mony$^{3}$, Soumya Punnathanam$^{3}$, Rahul Thapa$^{3}$, Nikesh Kotecha, PhD$^{3}$, Nigam H. Shah, MBBS, PhD$^{4}$, Jonathan H. Chen, MD, PhD$^{4}$}

\institutes{
    $^1$Department of Biomedical Data Science, Stanford, California, USA \\
    $^2$Stanford Children's Health, Palo Alto, California, USA \\
    $^3$Stanford Health Care, Palo Alto, California, USA \\
    $^4$Center for Biomedical Informatics Research, Stanford, California, USA \\
    $^*$ denotes equal contribution
}

\maketitle

\textit{Machine learning (ML) applications in healthcare are extensively researched, but successful translations to the bedside are scant.  Healthcare institutions are establishing frameworks to govern and promote the implementation of accurate, actionable and reliable models that integrate with clinical workflow. Such governance frameworks require an accompanying technical framework to deploy models in a resource efficient manner. Here we present DEPLOYR, a technical framework for enabling real-time deployment and monitoring of researcher created clinical ML models into a widely used electronic medical record (EMR) system. We discuss core functionality and design decisions, including mechanisms to trigger inference based on actions within EMR software, modules that collect real-time data to make inferences, mechanisms that close-the-loop by displaying inferences back to end-users within their workflow, monitoring modules that track performance of deployed models over time, silent deployment capabilities, and mechanisms to prospectively evaluate a deployed model's impact. We demonstrate the use of DEPLOYR by silently deploying and prospectively evaluating twelve ML models triggered by clinician button-clicks in Stanford Health Care's production instance of Epic. Our study highlights the need and feasibility for such silent deployment, because prospectively measured performance varies from retrospective estimates. By describing DEPLOYR, we aim to inform ML deployment best practices and help bridge the model implementation gap.}


\section*{Introduction}
Access to real-world data streams like electronic medical records (EMRs) has accelerated the promise of machine learning (ML) applications in healthcare. There exist over 250,000 peer-reviewed articles related to risk-stratification models alone, many of which have been published in the past ten years \cite{challener2019proliferation,guo2021systematic}.  Despite the hype, a sizeable gap separates ML models in peer-reviewed research articles from those actually impacting clinical care \cite{chen2017machine, matheny2019artificial}. This implementation gap leaves most published clinical ML applications lost in the ``model graveyard'' \cite{seneviratne2020bridging}.  

The gap between research and implementation demonstrates that strong predictive performance alone is not sufficient for feasible and worthwhile translation of a clinical ML model to the bedside.  Beyond demonstrating predictive accuracy, model champions must articulate concrete actions that can be taken as a result of inferences (model outputs) \cite{callahan2017machine}. Proposed actions must incur utility, some measurably positive effect on a clinical outcome of interest benefiting the deployment population as a whole \cite{shah2019making,ko2021improving, jung2021framework}. Positive impact on average, however, does not imply positive impact for all.  Successful ML applications to healthcare must satisfy fairness principles, ensuring performance and accrued utility is not unfavorable across certain patient populations, particularly those traditionally under-served \cite{hernandez2020minimar, lu2022considerations, char2018implementing}. Institutions leading the charge in the translation of clinical ML applications are establishing governance frameworks to ensure models are safe, reliable and useful throughout their deployment life-cycles \cite{armitage_2022, bedoya2022framework, reddy2020governance, wiens2019no}.

Even if the above are addressed and a go decision is made, ML applications must overcome technical feasibility hurdles related to their deployment. Traditionally, if such technical implementation were possible at all within the context of existing vendor and legacy infrastructure, massive overhead was required to establish and maintain computational frameworks enabling the deployment of ML models into clinical workflow \cite{sendak2017barriers, sculley2015hidden, morse2020estimate}.  Some early adopting institutions rely on frameworks native to their EMR vendor (e.g Epic Cognitive Computing) to deploy ML applications \cite{siwicki_2021}. These frameworks can reduce technical overhead and facilitate model sharing across health systems, but deployment of custom (researcher developed) ML models trained using an institution's clinical data-warehouse remains difficult. Health institutions that have successfully developed custom deployment frameworks have largely done so through tight collaborations of data scientists who build and validate models and hospital information technology (IT) teams who know the ins-and-outs of their institution's EMR \cite{kashyap2021survey}.  These combinations of expertise are valuable but rare to find. Successful implementations allow flexibility out-of-the-box vendor solutions do not, but design details largely remain internal knowledge, leaving each health system to either re-invent the wheel or be left behind. This has led to a sparse and eclectic set of solutions with limited academic discourse of best practices.  

Through a collaboration of data scientists at the Stanford School of Medicine, and IT persona at both Stanford Health Care and Stanford Children's Health, our objective was to develop DEPLOYR — a technical framework for deploying researcher developed clinical ML applications directly into our EMR's production environment. In this article, we detail the concrete mechanisms DEPLOYR enables and the design decisions made during its formation to provide a blueprint for development and best practices for execution. The key functions reviewed here include real-time closed-loop model trigger, data retrieval, inference, user interface integration, and continuous monitoring. We then demonstrate the capability and importance of prospective evaluations that extend beyond traditional retrospective analyses by silently deploying twelve ML models triggered by clinician button-clicks in Stanford Health Care's production instance of Epic.

\section*{Methods}
Effective deployments of clinical ML models require implementation of several core mechanisms. Here we detail these functions and discuss design decisions considered in our framework's development. Core functions include data sourcing (at train and inference time), mechanisms that trigger model inference in production, and mechanisms that integrate inferences and recommendations back to end-users.  We detail our implementation of a monitoring module that tracks performance of deployed models over time, mechanisms that enable silent trial deployments, and prospective evaluation of model impact.  Our deployment framework (DEPLOYR) enables these mechanisms by leveraging capabilities native to Stanford Health Care's EMR (Epic Systems) in conjunction with three distinct software applications: \textit{DEPLOYR-dev} (a python package used for model development and validation), \textit{DEPLOYR-serve} (a python Azure Function application used to expose trained models as APIs), and \textit{DEPLOYR-dash} (a dashboard implemented using the streamlit python package) \cite{microsoftAzureFunctions, streamlit}. Figure 1 provides a system level overview of the steps involved in a DEPLOYR enabled deployment.

\subsection*{Data sources}
Any ML deployment framework requires decisions on how to source data at both train and inference time. To effectively translate research models, the framework should expect training data to be sourced from research grade clinical data-warehouses. At inference time, many ML use cases require access to real-time data streams \cite{dash2019big, wiens2019no,corbin2022personalized,xu2019prevalence}.  Since clinical data warehouses are typically several transformations removed from real-time data streams, data mappings must be implemented between the two sources to ensure models receive the same data elements in training and deployment environments.

\paragraph{Training data source}
DEPLOYR uses data from Stanford's clinical data warehouse (STARR) for training \cite{datta2020new}. STARR contains de-identified EMR data from over 2.4 million unique patients spanning 2009-2021 who have visited Stanford Hospital (academic medical center in Palo Alto, California), ValleyCare hospital (community hospital in Pleasanton, California) and Stanford University Healthcare Alliance affiliated ambulatory clinics. STARR tables are hosted on Google BigQuery through a secured Google Cloud instance safe for protected health information (PHI) and maintained by the Stanford School of Medicine \cite{google}. STARR data relevant for model construction include patient demographics, comorbidities, procedures, medications, laboratory results, and vital signs.  Use of STARR data was approved by the institutional review board of the Stanford University School of Medicine.

\begin{figure}[t!]
\centering
\includegraphics[scale=0.60]{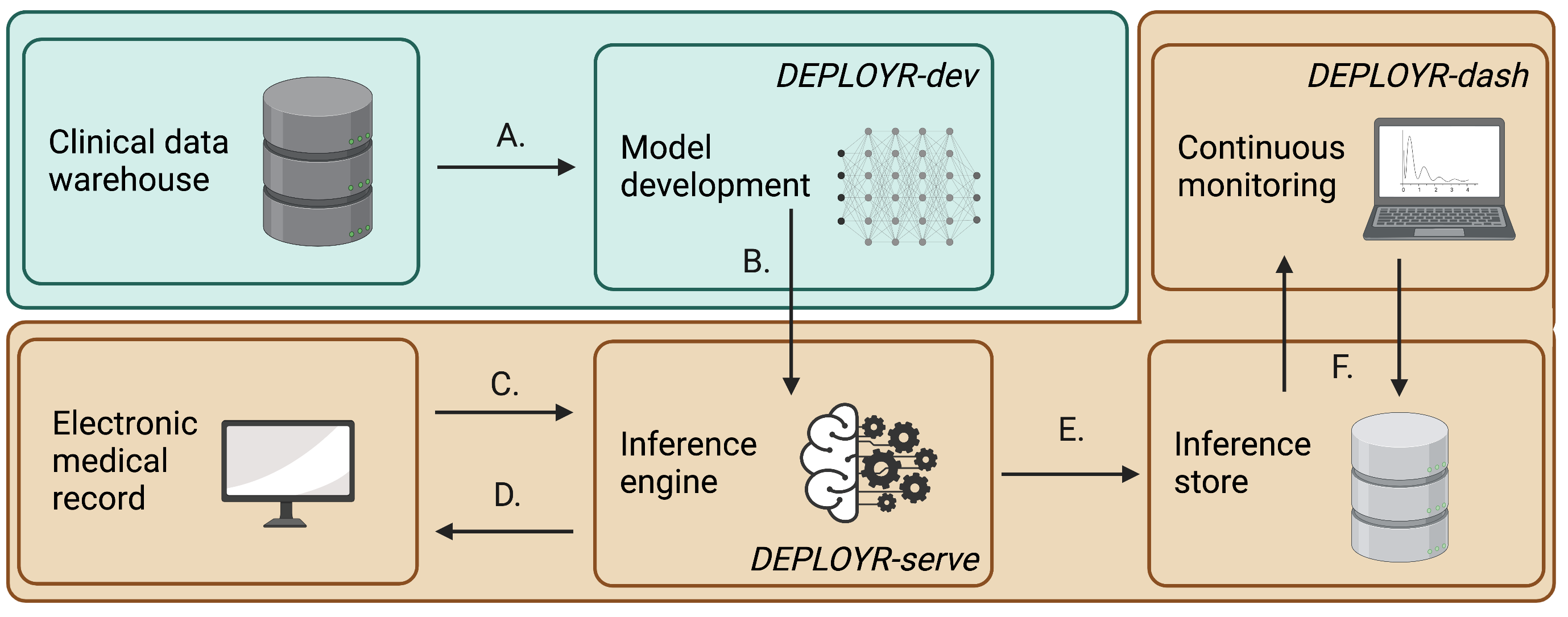}
\caption{Summary of a DEPLOYR enabled model deployment. Blue shading indicates infrastructure operated by the Stanford School of Medicine (academic research). Orange shading indicates infrastructure operated by Stanford Health Care (clinical operations). A. De-identified EMR data is sourced from Stanford's clinical data warehouse (STARR), a model is developed and retrospectively validated using the \textit{DEPLOYR-dev} python package.  B. The model is deployed to the inference engine, and exposed as a REST API using \textit{DEPLOYR-serve}. C. Inference is triggered, spawning an HTTPS request from the EMR directed at the exposed model. D. The request results in the collection of a feature vector from the EMR's transactional database using REST (both FHIR and EMR specific) APIs maintained by the EMR vendor. Inference is performed on the real-time retrieved data, and routed back to the EMR closing the loop with end-users and integrating into workflow. E. Inferences and relevant metadata are additionally saved to the inference store, (F.) and consumed by monitoring software (\textit{DEPLOYR-dash}) that continuously tracks model performance via dashboard. REST = Representation State Transfer. HTTPS = Hypertext Transfer Protocol Secure. FHIR = Fast Healthcare Interoperability Resources.}
\label{fig1}
\end{figure}

\paragraph{Inference data source}
In production, we source data directly from the EMR's transactional database, Epic Chronicles, which contains real-time patient data \cite{krall1995acceptance}. We access Epic Chronicles using Epic and FHIR (Fast Healthcare Interoperability Resources) REST APIs documented in Epic's App Orchard. \cite{bender2013hl7, barker2021ecosystem}. Credentials enabling authentication to these APIs were provisioned upon registration of a back-end application to Epic's App Orchard. HTTPS requests directed at these APIs enable real-time access to commonly used clinical ML features, including patient demographics, comorbidities, procedures, medications, laboratory results and vital signs.  Use of this data source falls under the umbrella of hospital operational work that is specifically scoped to quality improvement.  Data sources and mappings are depicted in Figure 2.

\subsection*{Model inference triggers}
Deployment frameworks require mechanisms to trigger model inference. The choice of a suitable inference triggering mechanism depends on how models expect to integrate with clinical workflow. Because triggering logic specifies a model's deployment population (determines which patients under which conditions receive an inference), it should be considered during cohort development to ensure the population in a researcher's retrospective test set matches what is seen in production. DEPLOYR supports two broad classes of inference triggering mechanisms — event and time based triggers. 


\paragraph{Event based triggers}
Event based triggers execute as a direct result of a clinical action. Examples include order entry for a laboratory diagnostic test, signature of a progress note, inpatient admission, or discharge.  We enable event based triggers by exposing deployed models as REST APIs that listen for inbound HTTPS requests from Epic.  Models are wrapped in custom python functions exposed as REST APIs using \textit{DEPLOYR-serve}, an Azure Function application deployed to Stanford Health Care's instance of Azure \cite{microsoftAzureFunctions}. HTTPS requests are spawned from the EMR through use of Epic Best Practice Advisories (BPAs), rules, and programming points configured to execute upon button clicks in Epic Hyperspace, Epic's graphical user-interface \cite{klatt2012effect, ahmed2022interacting}.  HTTPS requests from Epic transmit patient identifiers to \textit{DEPLOYR-serve} functions enabling feature vector collection specific to the patients for whom inference was triggered. \textit{DEPLOYR-serve} functions are exposed only to traffic within Stanford Health Care's internal network.

\paragraph{Time based triggers}
Time based triggers initiate inference at preset time intervals on batches of patients at once. Example applications could include models designed to monitor patients for signs of deterioration, sepsis, or acute kidney injury by performing inference on collections of patients every fifteen minutes. \cite{ye2019real, nemati2018interpretable, saqib2018early, tomavsev2019clinically}.  \textit{DEPLOYR-serve} supports time based triggering through the use of Azure Function timer triggers, which we configure using cron logic \cite{microsoftAzureFunctions, keller1999take}. Unlike event based triggers that execute inference for specific patients, time based triggers initiate inference on batches. Because time based triggers do not originate from Epic, ML models that use them must first identify for which patients inference should be performed. \textit{DEPLOYR-serve} functions fetch this information using Epic APIs that collect batches of patients identifiers, for example, within specific hospital units.  Event and time based triggers are summarized in Figure 3.

\begin{figure}[t!]
\centering
\includegraphics[scale=0.65]{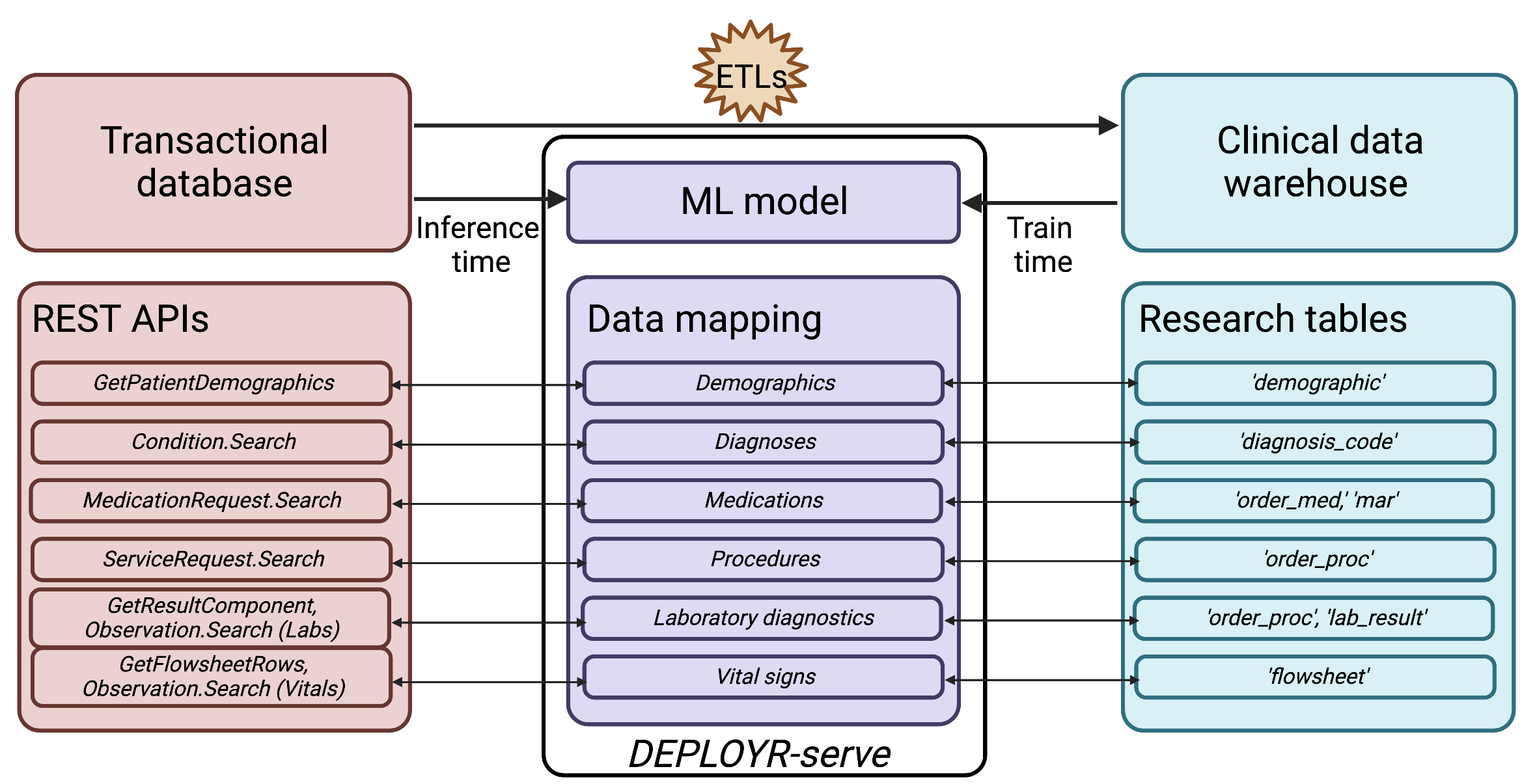}
\caption{ML models are trained using data sourced from the Stanford's clinical data warehouse (STARR). In production, real-time data is sourced from the EMR's transactional database (Epic Chronicles) through Epic and FHIR REST APIs.  STARR data are several ETLs (extract, transform, loads) removed from the transactional database. Data mapping is necessary at inference time to ensure features seen during training match features seen in production.  Mappings and inferences are invoked in \textit{DEPLOYR-serve}.}
\label{fig2}
\end{figure}

\subsection*{Directing model outputs}
Deployment frameworks require mechanisms to direct model inferences and recommendations back to end-users, closing the loop. Although it is possible to communicate inferences via third-party web or mobile applications, We integrate inferences directly into the EMR to reduce technological overhead and best fit with clinical workflow. Inference integration with Epic is possible through several distinct mechanisms. Inference integration is implemented in \textit{DEPLOYR-serve}.

\paragraph{Passive integration}
Model inferences can be written back to the EMR without interrupting clinical workflow. An example includes directing inferences to Epic displayed as external model score columns visible in inpatient lists and outpatient schedules. Score columns can be configured to display predicted probabilities, binary flags, feature values and their contributions (e.g., Shapley values) to promote inference level interpretability \cite{NIPS2017_7062}. Additionally, inferences can be written to flowsheet rows and smart data values, which can be used to trigger downstream clinical decision support \cite{flynn2022tracking}.  Inferences and suggested interventions can also be directed at clinicians using Epic's in-basket messaging platform.  Inferences are routed back to the EMR by way of HTTPS requests directed at APIs exposed and documented by Epic.  Writing to external model score columns requires the Epic Cognitive Computing API \cite{siwicki_2021}. Writing to smart data values, flowsheet rows and sending in-basket messages does not. 

\paragraph{Active integration}
Some applications will better integrate with workflow through interruptive alerts in the clinical work-stream. Such active integration is possible through use of typical EMR alerts. Specifically, we use Epic BPA (Best Practice Advisory) web services.  Epic supports two styles of web services: classic CDA (clinical document architecture) and CDS web-hooks \cite{goldberg2016use}. DEPLOYR currently uses classic CDA web services, which direct HTTPS requests at \textit{DEPLOYR-serve} functions and await XML (Extensible Markup Language) responses \cite{bloomfield2017opening}. Data packaged in the XML response is then displayed as an alert in the EMR's user interface.  We show mock-ups of EMR integration capabilities in Figure 4.

\begin{figure}[t!]
\centering
\includegraphics[scale=0.4]{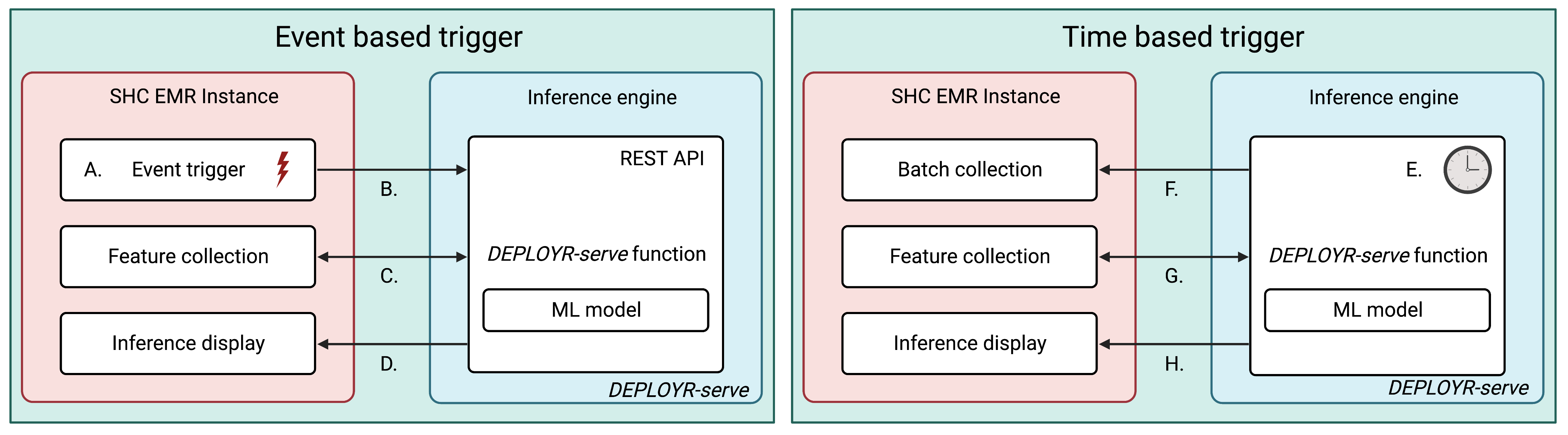}
\caption{DEPLOYR triggering mechanisms. Models deployed with event based triggering logic are exposed as REST APIs on the inference engine using a python Azure Function application (\textit{DEPLOYR-serve}). An event (A) in the EMR (e.g clinician button-click initiating a laboratory order) transmits an HTTPS request (B) directed at the exposed \textit{DEPLOYR-serve} function, which wraps an ML model. The function transmits HTTPS requests (C) to REST APIs documented in Epic's App Orchard to collect a feature vector, performs model inference, and directs the inference and resulting clinical decision support via HTTPS request (D) back into the EMR to interface with end-users.  Models deployed with time based triggering logic perform inference at set intervals (E) through use of Azure Function timer triggers. Every time interval (e.g 15 minutes), a \textit{DEPLOYR-serve} function transmits HTTPS requests (F) to REST APIs to retrieve a batch of patient identifiers for whom inference should be made.  Feature vectors are collected for the batch of patients (G), and inferences are transmitted back into the EMR (H). SHC = Stanford Health Care.}
\label{fig3}
\end{figure}

\begin{figure}[t!]
\centering
\includegraphics[scale=0.39]{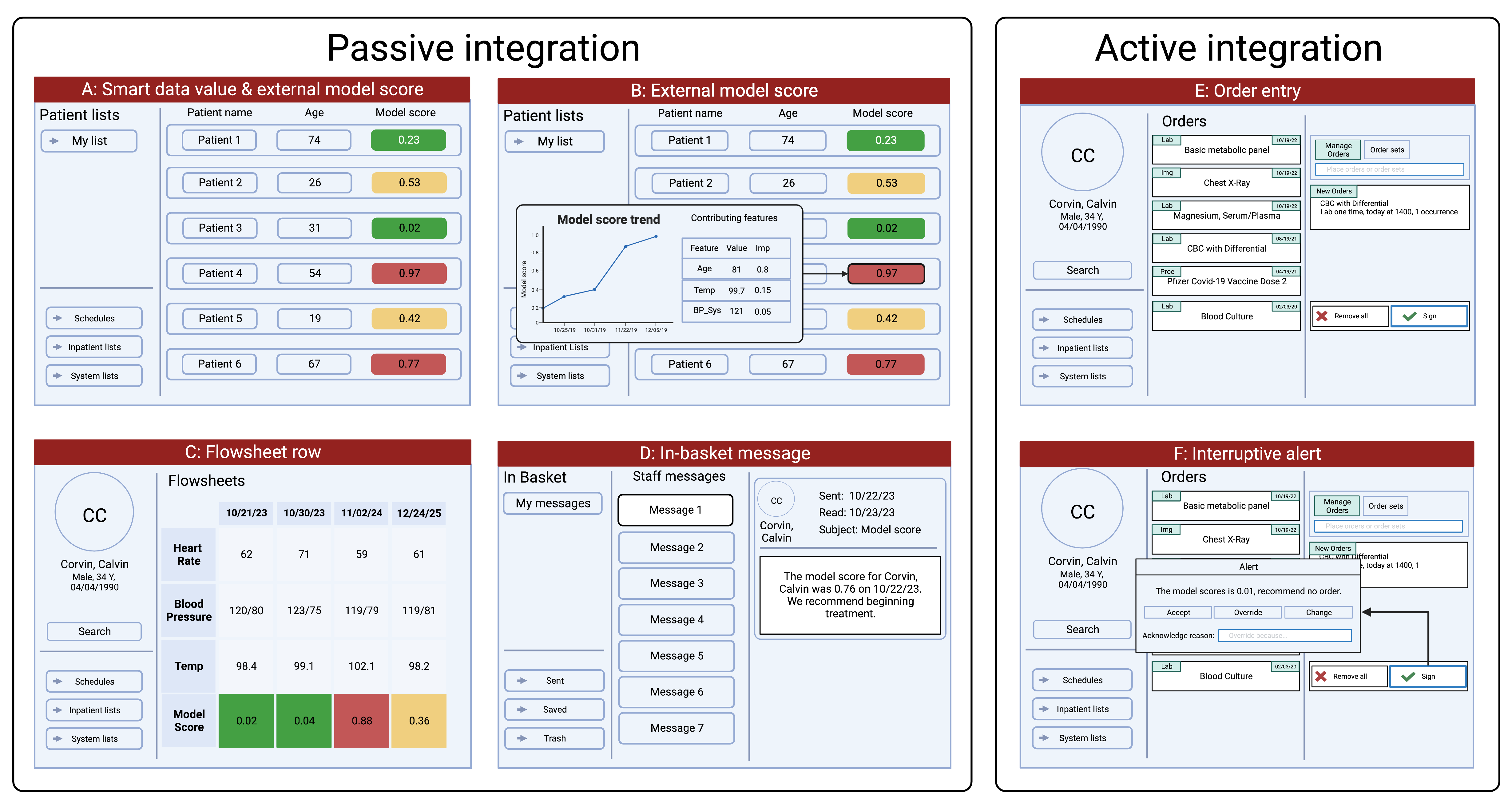}
\caption{Mock-up frames depicting the EMR user-interface and mechanisms in which inferences (model outputs) can be displayed to end-users. Inferences can be directed back into the EMR passively, without interrupting clinical workflow. They can be written as smart data values or as external model score columns displayed in inpatient lists and outpatient schedules (A).  Hovering the mouse over an external model score column value in a patient list displays a pop-up window (B) depicting the trend of the model score over time, feature values, and contribution of those features to the resulting inference.  Model inferences can be written to flowsheet rows (C) and visualized over time in conjunction with other vital sign data (e.g. heart rate, blood pressure, temperature). Inferences and suggested interventions can be directed as in-basket messages (D) to specific providers.  Additionally, inferences can be integrated with the EMR actively through use of interruptive alerts (E, F) that trigger as a result of button-clicks (e.g. signature of a laboratory order) in the user-interface. Inference integration is implemented in \textit{DEPLOYR-serve}.}
\label{fig4}
\end{figure}

\subsection*{Continuous performance monitoring}
Once ML models are deployed, they must be monitored to ensure continued reliability in production \cite{feng2022clinical, schroder2022monitoring, klaise2020monitoring}.  All ML models are subject to potential performance decay over time due to changes in the underlying data distribution. Examples including covariate shift (changes in the distribution over features), label shift (changes in distribution over labels) and concept shift (changes in the distribution of labels conditioned on features) \cite{jung2015implications, guo2021systematic}.  Suitable deployment infrastructure requires monitoring software to track model performance metrics, as well as feature and label distributions throughout the model's life cycle. 

\paragraph{Extracting labels in production}
To track ML performance in production, software must be implemented to collect labels. DEPLOYR uses \textit{LabelExtactors} to perform this task.  
\textit{LabelExtractors} are implemented in \textit{DEPLOYR-serve} and are specific to a particular deployed model.  They execute at user-defined frequencies using cron logic.  At inference time, in addition to directing model outputs to Epic, we package inferences with relevant metadata (identifiers, timestamps, features) in inference packets (JSON objects) and save them to the inference store, an Azure Cosmos database maintained by Stanford Health Care. \textit{LabelExtactors} consume inference packets from the inference store and pair to them their corresponding labels (once observable) using Epic and FHIR APIs. A model tasked with predicting unplanned (30-day readmission) might produce inference packets that include the patient's FHIR identifier and discharge time (inference time) along with the inference itself. The corresponding \textit{LabelExtractor} would then use the FHIR identifier and discharge time to make HTTPS requests directed at relevant APIs, determining whether the patient was readmitted to the hospital within 30 days of the produced inference.


\paragraph{Tracking model performance}
After a \textit{LabelExtractor} has paired a set of inference packets to their corresponding labels, standard model performance metrics can be estimated with the prospectively collected data.  When monitoring binary classifiers, standard performance metrics may include threshold dependent measures like accuracy, sensitivity (recall), specificity, positive predictive value (precision) along with threshold independent metrics like area under the receiver operator characteristics curve (AUROC) and average precision.  Standard performance plots that track discrimination ability (ROC and precision-recall curves) and calibration (calibration curves) can be constructed. In addition to full population estimates, these metrics can be continuously tracked over patient subgroups, particularly those belonging to protected demographic classes.  Beyond classic ML metrics, measures of model usefulness such as net benefit and expected utility can be tracked to ensure model use is yielding more good than harm. \cite{vickers2016net, shah2019making}

\paragraph{Tracking distribution shifts in features and labels}
Machine learning performance decay over-time is often attributable to changes in the underlying data distribution.  Beyond calculating prospective performance metrics, DEPLOYR monitoring infrastructure tracks statistics (e.g mean of data elements across groups of patients) that describe the distributions of features, labels and predictions over time. When deviations in model performance arise, corresponding changes to underlying feature and label distributions can be the reason.  Both model performance and distributional statistics are displayed to a dashboard created using the streamlit python package, which we show in Figure 5 \cite{streamlit}.  

\begin{figure}[t!]
\centering
\includegraphics[scale=0.40]{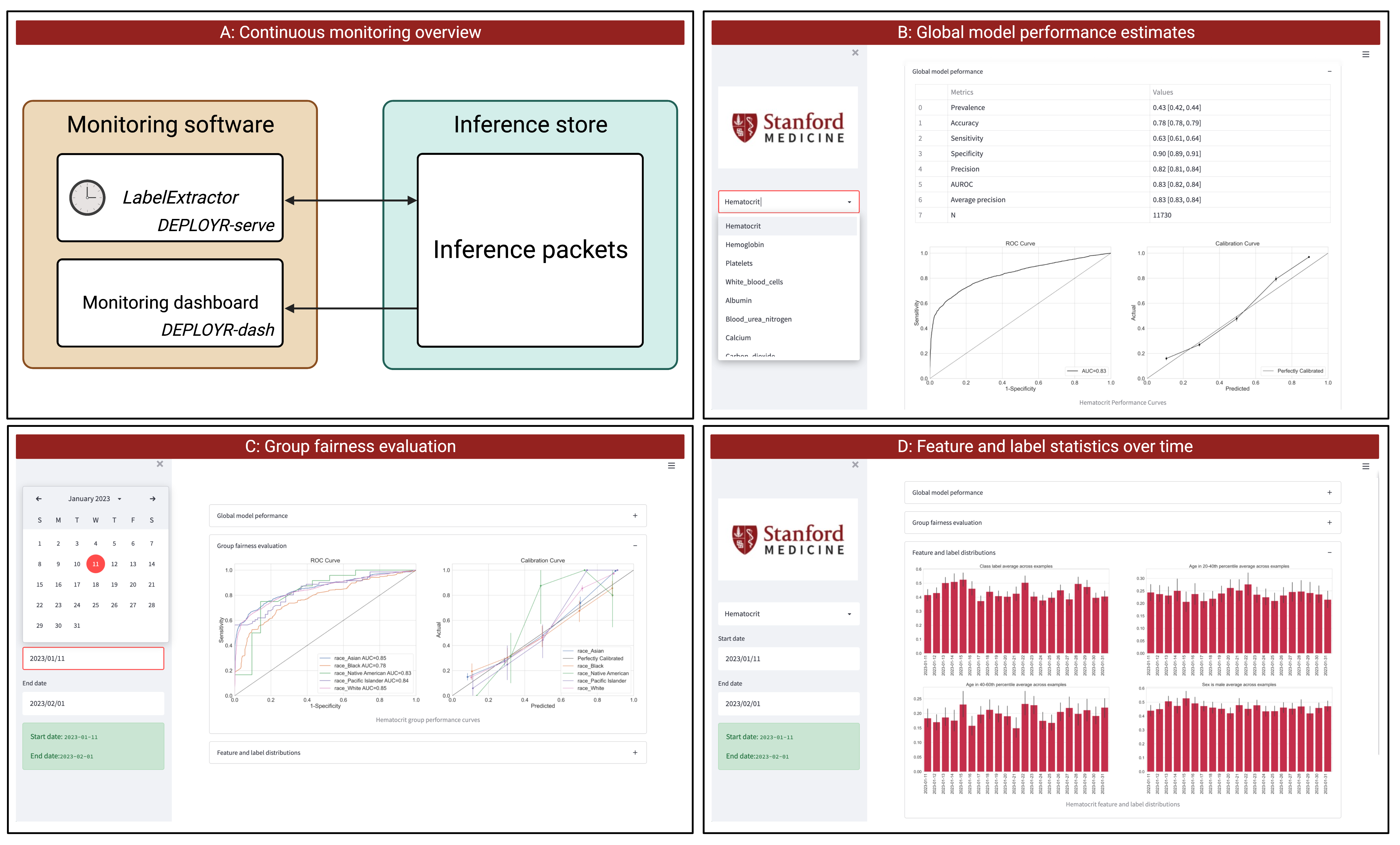}
\caption{DEPLOYR performance monitoring. Panel A depicts the flow of data  from monitoring software to the inference store. A \textit{LabelExtactor} is implemented in \textit{DEPLOYR-serve} as an Azure Function timer trigger. When executed it collects inference packets and pairs them with their corresponding labels. Inference packets paired with labels are consumed by a monitoring dashboard, implemented in \textit{DEPLOYR-dash}.  Panel B shows a screenshot of the dashboard displaying global model performance for the user-specified model. Panel C shows a group fairness evaluation across protected demographic classes for the user-specified time window. Panel D shows feature and label distribution statistics collected and tracked over time.}
\label{fig5}
\end{figure}


\subsection*{Enabling silent deployment}
Suitable deployment infrastructure requires the ability to silently deploy models prospectively to ensure reliability in production before ``loud'' deployment when inferences are displayed to end-users \cite{tonekaboni2022validate,bedoya2022framework}. Silent trials allow a data science team responsible for model deployment to ensure data pipes have been appropriately linked, especially useful when training and deployment data sources differ.  Additionally, silent deployment can uncover faulty cohort design, for instance when a retrospective cohort was generated using exclusion criteria observable only after inference time, common when generating ML cohorts using case-control study design \cite{krautenbacher2017correcting, reps2021design}. DEPLOYR event and time based triggers can be configured to execute in the background, enabling silent trials. 

\subsection*{Enabling prospective evaluation of impact}
The ultimate evaluation of healthcare machine learning impact is not in a prospective ROC curve or simulation of clinical utility, but estimation of the causal effect of the model's implementation on clinical and operational outcomes. In some use-cases, a model's inference and corresponding recommendation will only be displayed to the end-user if a predicted risk exceeds some threshold value.  In these situations, regression discontinuity designs may be suitable to estimate the local treatment effect of the model's implementation for patients with predicted risk near the threshold \cite{david2019effect}. In other use-cases when a model's inference is displayed irrespective of the value it takes, or when an average treatment effect on the deployment population at large is desired, randomized study designs may be required \cite{nemati2022randomized, escobar2020automated}.  DEPLOYR supports functionality to inject randomization into the inference-directing mechanism to enable prospective determination of a model's impact.  
	
\section*{Results}
We used the DEPLOYR framework to silently deploy twelve binary ML classifiers that predict laboratory diagnostic results, building off of previous retrospective analyses of models designed to reduce wasteful laboratory utilization \cite{xu2019prevalence, aikens2019machine, rabbani2023targeting, kim2021machine}. Models were trained with de-identified EMR data from Stanford's clinical data warehouse (STARR) using \textit{DEPLOYR-dev}, and exposed as REST APIs using \textit{DEPLOYR-serve}. Epic BPAs were configured to spawn HTTPS requests directed at our deployed models upon order signature of the diagnostic test whose result each model aimed to predict — instances of event-based triggering mechanisms.  

\begin{table}[t!]
\centering
\small
\caption{Demographic breakdown of retrospective and prospective cohorts.  Retrospective cohorts were sourced from Stanford's clinical data warehouse (STARR).  Prospective cohorts were collected in real-time through Epic Chronicles as diagnostic orders triggered model inferences between January 11 and February 15, 2023. CBC=Complete blood count with differential.}
\begin{tabular}{|ll|lll|lll|}
\hline
\multicolumn{2}{|l|}{\textbf{Demographic breakdown}}                                     & \multicolumn{3}{l|}{\textbf{Retrospective cohorts}}                                              & \multicolumn{3}{l|}{\textbf{Prospective cohorts}}                                                \\ \hline
\multicolumn{2}{|l|}{\textbf{Diagnostic}}                                                & \multicolumn{1}{l|}{\textit{CBC}} & \multicolumn{1}{l|}{\textit{Magnesium}} & \textit{Metabolic} & \multicolumn{1}{l|}{\textit{CBC}} & \multicolumn{1}{l|}{\textit{Magnesium}} & \textit{Metabolic} \\ \hline
\multicolumn{2}{|l|}{\textbf{N unique patients}}                                                         & \multicolumn{1}{l|}{13362}        & \multicolumn{1}{l|}{11771}              & 13410              & \multicolumn{1}{l|}{18982}        & \multicolumn{1}{l|}{5234}               & 16441              \\ \hline
\multicolumn{2}{|l|}{\textbf{Age, mean (SD)}}                                            & \multicolumn{1}{l|}{51.4 (23.7)}  & \multicolumn{1}{l|}{53.1 (24.2)}        & 54.6 (21.4)        & \multicolumn{1}{l|}{55.7 (21.4)}  & \multicolumn{1}{l|}{62.1 (17.8)}        & 55.4 (21.1)        \\ \hline
\multicolumn{1}{|l|}{\multirow{3}{*}{\textbf{Sex, n (\%)}}}  & \textit{Female}           & \multicolumn{1}{l|}{7103 (53.2)}  & \multicolumn{1}{l|}{5432 (46.1)}        & 6923 (51.6)        & \multicolumn{1}{l|}{10092 (53.2)} & \multicolumn{1}{l|}{2425 (46.3)}        & 8711 (53.0)        \\ \cline{2-8} 
\multicolumn{1}{|l|}{}                                       & \textit{Male}             & \multicolumn{1}{l|}{6259 (46.8)}  & \multicolumn{1}{l|}{6338 (53.8)}        & 6485 (48.4)        & \multicolumn{1}{l|}{8884 (46.8)}  & \multicolumn{1}{l|}{2808 (53.6)}        & 7724 (47.0)        \\ \cline{2-8} 
\multicolumn{1}{|l|}{}                                       & \textit{Unknown}          & \multicolumn{1}{l|}{0 (0.0)}      & \multicolumn{1}{l|}{1 (0.0)}            & 2 (0.0)            & \multicolumn{1}{l|}{6 (0.0)}      & \multicolumn{1}{l|}{1 (0.0)}            & 6 (0.0)            \\ \hline
\multicolumn{1}{|l|}{\multirow{7}{*}{\textbf{Race, n (\%)}}} & \textit{White}            & \multicolumn{1}{l|}{6888 (51.5)}  & \multicolumn{1}{l|}{6038 (51.3)}        & 6915 (51.6)        & \multicolumn{1}{l|}{9009 (47.5)}  & \multicolumn{1}{l|}{2726 (52.1)}        & 7636 (46.4)        \\ \cline{2-8} 
\multicolumn{1}{|l|}{}                                       & \textit{Other}            & \multicolumn{1}{l|}{2870 (21.5)}  & \multicolumn{1}{l|}{2899 (24.6)}        & 2680 (20.0)        & \multicolumn{1}{l|}{4545 (23.9)}  & \multicolumn{1}{l|}{1067 (20.4)}        & 4103 (25.0)        \\ \cline{2-8} 
\multicolumn{1}{|l|}{}                                       & \textit{Asian}            & \multicolumn{1}{l|}{2444 (18.3)}  & \multicolumn{1}{l|}{1739 (14.8)}        & 2600 (19.4)        & \multicolumn{1}{l|}{3612 (19.0)}  & \multicolumn{1}{l|}{928 (17.7)}         & 3125 (19.0)        \\ \cline{2-8} 
\multicolumn{1}{|l|}{}                                       & \textit{Black}            & \multicolumn{1}{l|}{557 (4.2)}    & \multicolumn{1}{l|}{587 (5.0)}          & 583 (4.3)          & \multicolumn{1}{l|}{1060 (5.6)}   & \multicolumn{1}{l|}{327 (6.2)}          & 884 (5.4)          \\ \cline{2-8} 
\multicolumn{1}{|l|}{}                                       & \textit{Unknown}          & \multicolumn{1}{l|}{391 (2.9)}    & \multicolumn{1}{l|}{238 (2.0)}          & 420 (3.1)          & \multicolumn{1}{l|}{382 (2.0)}    & \multicolumn{1}{l|}{56 (1.1)}           & 353 (2.1)          \\ \cline{2-8} 
\multicolumn{1}{|l|}{}                                       & \textit{Pacific Islander} & \multicolumn{1}{l|}{168 (1.3)}    & \multicolumn{1}{l|}{212 (1.8)}          & 171 (1.3)          & \multicolumn{1}{l|}{294 (1.5)}    & \multicolumn{1}{l|}{98 (1.9)}           & 261 (1.6)          \\ \cline{2-8} 
\multicolumn{1}{|l|}{}                                       & \textit{Native American}  & \multicolumn{1}{l|}{44 (0.3)}     & \multicolumn{1}{l|}{58 (0.5)}           & 41 (0.3)           & \multicolumn{1}{l|}{80 (0.4)}     & \multicolumn{1}{l|}{32 (0.6)}           & 79 (0.5)           \\ \hline
\end{tabular}
\end{table}

\subsection*{Cohort and prediction task definitions}
We constructed three retrospective cohorts with STARR data to develop and validate twelve binary ML classifiers.  Cohorts were specific to the laboratory diagnostic exam whose results our models predict. The unit of observation was an order for a diagnostic exam. One cohort included orders for complete blood count (CBC) with differential diagnostics. Another included orders of metabolic panels. A third included orders for magnesium diagnostics. CBC and metabolic panel diagnostics result in several components, whereas magnesium tests yield a single result.  Consistent with prior work, we trained independent binary classifiers per component to, at order time, predict whether the test result would fall outside the clinical laboratory defined normal reference range \cite{rabbani2023targeting, xu2019prevalence}.  The CBC cohort was used to train four models that separately predicted hematocrit, hemogloblin, white blood cell, and platelet results. The metabolic panel cohort was used to train seven models that predicted albumin, blood urea nitrogen, calcium, carbon dioxide, creatinine, potassium, and sodium results. We trained one model to predict magnesium results. All three retrospective cohorts contained data from 2015 to 2021, with two thousand diagnostic tests sampled randomly per year for a total of 14,000 orders per task.  Corresponding prospective cohorts were collected in real-time during our silent deployment trial — spanning the dates January 11 to February 15, 2023.  Prospective cohorts were not used to further tune our models, but rather served as additional test sets for model evaluation collected directly in the production environment.  Demographic characteristics of patients included in both retrospective and prospective cohorts are summarized in Table 1. 


\subsection*{Model training and evaluation}
Random forest classifiers were fit for each task.  Features included patient demographics, diagnosis codes mentioned on the problem list, medication orders, and prior lab results. Features were represented as counts \cite{rajkomar2018scalable}, with numerical data first binned into quintiles based on distributions observed in the training set — as detailed in Supplemental Note 1. All twelve ML models were evaluated using retrospective and prospective test sets. We measured model discrimination ability by estimating the area under the receiver operating characteristics curve (AUROC). 95\% confidence intervals were estimated by bootstrapping the corresponding test set 1000 times. We report these estimates along with the prevalence of the positive class (an abnormal diagnostic result) in Table 2. Additionally, we constructed retrospective and prospective receiver operating characteristics (ROC) curves, precision recall (PR) curves, and calibration plots — shown in the Supplemental Figures 6-8. Beyond tracking global performance measures, we estimate model performance across patient subgroups stratified by protected demographic classes in both retrospective and prospective test sets.  These results are shown in the Supplemental Tables 3-14. 

\begin{table}[t!]
\centering
\small
\caption{Model performance estimates on retrospective and prospectively collected test sets for all twelve models. AUROC=area under the receiver operating characteristics curve.}
\begin{tabular}{|ll|ll|ll|}
\hline
\multicolumn{2}{|l|}{\textbf{Model performance}}                                                                                               & \multicolumn{2}{l|}{\textbf{Positive (abnormal) prevalence}}        & \multicolumn{2}{l|}{\textbf{AUROC}}                                 \\ \hline
\multicolumn{1}{|l|}{\textbf{Diagnostic}}                                                                       & \textbf{Component}           & \multicolumn{1}{l|}{\textit{Retrospective}} & \textit{Prospective}  & \multicolumn{1}{l|}{\textit{Retrospective}} & \textit{Prospective}  \\ \hline
\multicolumn{1}{|l|}{\multirow{4}{*}{\textbf{\begin{tabular}[c]{@{}l@{}}CBC with\\ differential\end{tabular}}}} & \textit{Hematrocrit}         & \multicolumn{1}{l|}{0.47 {[}0.45, 0.49{]}}  & 0.43 {[}0.42, 0.44{]} & \multicolumn{1}{l|}{0.86 {[}0.85, 0.88{]}}  & 0.83 {[}0.83, 0.84{]} \\ \cline{2-6} 
\multicolumn{1}{|l|}{}                                                                                          & \textit{Hemoglobin}          & \multicolumn{1}{l|}{0.50 {[}0.48, 0.52{]}}  & 0.47 {[}0.46, 0.47{]} & \multicolumn{1}{l|}{0.88 {[}0.86, 0.89{]}}  & 0.83 {[}0.83, 0.84{]} \\ \cline{2-6} 
\multicolumn{1}{|l|}{}                                                                                          & \textit{Platelets}           & \multicolumn{1}{l|}{0.26 {[}0.24, 0.28{]}}  & 0.21 {[}0.21, 0.22{]} & \multicolumn{1}{l|}{0.79 {[}0.77, 0.82{]}}  & 0.77 {[}0.76, 0.78{]} \\ \cline{2-6} 
\multicolumn{1}{|l|}{}                                                                                          & \textit{White blood cell}    & \multicolumn{1}{l|}{0.29 {[}0.27, 0.31{]}}  & 0.27 {[}0.27, 0.28{]} & \multicolumn{1}{l|}{0.76 {[}0.74, 0.79{]}}  & 0.69 {[}0.68, 0.70{]} \\ \hline
\multicolumn{1}{|l|}{\multirow{7}{*}{\textbf{\begin{tabular}[c]{@{}l@{}}Metabolic\\ panel\end{tabular}}}}       & \textit{Albumin}             & \multicolumn{1}{l|}{0.20 {[}0.18, 0.21{]}}  & 0.21 {[}0.20, 0.21{]} & \multicolumn{1}{l|}{0.88 {[}0.86, 0.91{]}}  & 0.85 {[}0.84, 0.86{]} \\ \cline{2-6} 
\multicolumn{1}{|l|}{}                                                                                          & \textit{Blood urea nitrogen} & \multicolumn{1}{l|}{0.22 {[}0.20, 0.24{]}}  & 0.24 {[}0.23, 0.25{]} & \multicolumn{1}{l|}{0.85 {[}0.83, 0.87{]}}  & 0.80 {[}0.79, 0.81{]} \\ \cline{2-6} 
\multicolumn{1}{|l|}{}                                                                                          & \textit{Calcium}             & \multicolumn{1}{l|}{0.11 {[}0.10, 0.13{]}}  & 0.11 {[}0.11, 0.12{]} & \multicolumn{1}{l|}{0.80 {[}0.76, 0.83{]}}  & 0.79 {[}0.78, 0.81{]} \\ \cline{2-6} 
\multicolumn{1}{|l|}{}                                                                                          & \textit{Carbon dioxide}      & \multicolumn{1}{l|}{0.16 {[}0.14, 0.17{]}}  & 0.20 {[}0.20, 0.21{]} & \multicolumn{1}{l|}{0.69 {[}0.66, 0.72{]}}  & 0.62 {[}0.61, 0.63{]} \\ \cline{2-6} 
\multicolumn{1}{|l|}{}                                                                                          & \textit{Creatinine}            & \multicolumn{1}{l|}{0.31 {[}0.29, 0.33{]}}  & 0.32 {[}0.31, 0.33{]} & \multicolumn{1}{l|}{0.78 {[}0.75, 0.80{]}}  & 0.75 {[}0.74, 0.76{]} \\ \cline{2-6} 
\multicolumn{1}{|l|}{}                                                                                          & \textit{Potassium}           & \multicolumn{1}{l|}{0.06 {[}0.05, 0.08{]}}  & 0.08 {[}0.08, 0.09{]} & \multicolumn{1}{l|}{0.67 {[}0.61, 0.72{]}}  & 0.60 {[}0.59, 0.62{]} \\ \cline{2-6} 
\multicolumn{1}{|l|}{}                                                                                          & \textit{Sodium}              & \multicolumn{1}{l|}{0.12 {[}0.10, 0.13{]}}  & 0.17 {[}0.16, 0.17{]} & \multicolumn{1}{l|}{0.79 {[}0.75, 0.82{]}}  & 0.71 {[}0.70, 0.72{]} \\ \hline
\multicolumn{1}{|l|}{\textbf{Magnesium}}                                                                        & \textit{Magnesium}           & \multicolumn{1}{l|}{0.15 {[}0.14, 0.17{]}}  & 0.14 {[}0.13, 0.15{]} & \multicolumn{1}{l|}{0.70 {[}0.67, 0.73{]}}  & 0.65 {[}0.63, 0.67{]} \\ \hline
\end{tabular}
\end{table}

\section*{Discussion}
In this study we develop and demonstrate DEPLOYR, a technical framework for moving ML models trained on EMR data from research to production.  DEPLOYR was developed through a collaboration of researchers within the Stanford School of Medicine and IT persona at Stanford Health Care and Stanford Children's Health.  We detail design configurations, and evaluate our framework by developing, silently deploying and prospectively evaluating twelve ML models into Stanford Health Care's production instance of Epic.

To facilitate translation of researcher developed models, data is sourced from Stanford's clinical data warehouse at train time. At inference time, data is sourced from our EMR's transactional database.  Though possible to perform inference using other sources (e.g analytic or enterprise data warehouses), model inferences would not incorporate up-to-date data. Examples of use cases requiring real-time date are plentiful, ranging from sepsis prediction and early warning scores, to medication and laboratory diagnostic recommender systems, to emergency department triage \cite{ye2019real, nemati2018interpretable, saqib2018early, tomavsev2019clinically, lewin2021predicting,corbin2022personalized,corbin2020personalized, eaneff2020case, yelin2019personal, kanjilal2020decision, nguyen2021developing, xu2019prevalence, rabbani2023targeting}.

Related to the need for real-time data is the need for flexible inference triggering mechanisms.  Event-based triggers allow inference to spawn from button-clicks within our EMR's user interface (Epic Hyperspace). Example use cases include ML models that suggest diagnostic tests upon a patient admission, suggest minimizing wasteful diagnostics upon order entry, and suggest diagnosis coding upon note signatures \cite{xu2019prevalence, chen2016orderrex, chen2014automated, ip2022data, kim2022can, bajor2017predicting, schafer2019umls}. In other ML applications, inference is better integrated into workflow when executed on batches of patients — for example a sepsis prediction model that monitors all patients in the ICU every fifteen minutes \cite{adams2022prospective}.  We enable these workflows through time based triggers. To close the loop, we integrate inferences directly into the EMR, eliminating the need for external user facing applications that can increase friction and slow adoption.

Continuous monitoring tools are essential to any deployment framework \cite{feng2022clinical}. \textit{LabelExtractors} pair inferences with their corresponding labels when observable, and inference-label pairs are consumed by software that estimate relevant ML performance measures and statistics that describe the distribution of features, labels and predictions over time. Statistically significant and clinically meaningful variations in performance can serve as indicators suggesting models be re-trained or decommissioned \cite{rabanser2019failing, davis2020detection, davis2019nonparametric}. When monitoring deployed ML applications, careful attention to feedback mechanisms from interventions administered as a result of model integration must be considered. Feedback mechanisms when ignored can cause drastic deviations between observed and actual performance \cite{lenert2019prognostic, perdomo2020performative, adam2020hidden}. Naively updating these models do more harm than good \cite{adam2022error}.  DEPLOYR supports injecting randomization into the inference routing mechanism, which can be used in tandem with traditional weighting estimators to recover true performance estimates in the presence of feedback \cite{corbin2022avoiding}. The ability to randomize is similarly critical for deliberate randomized controlled trial evaluations of ML model impact on clinical outcomes \cite{nemati2022randomized, escobar2020automated}.
 
Silent trials are necessary before ML applications are integrated into clinical work-streams \cite{bedoya2022framework}. Deployment frameworks require mechanisms to trigger inference using production data without interacting with clinical work streams. As demonstrated by our silent trial, ML performance estimated on clinical data warehouse derived retrospective test sets can vary from prospectively observed performance in production.  Across our twelve laboratory predictions tasks, AUROC in our prospective test sets were generally several percentage points lower than what is seen retrospectively. Poorer performance could be attributable to data drift, data-elements appearing to be accessible at inference time in the clinical data-warehouse not actually being accessible in production, and imperfect mappings between training and inference data sources. Due to these deviations, we recommend using performance measures estimated from prospectively collected test sets during silent trials to make final go decisions when deploying clinical ML models. 

Limitations to this study include that, while the DEPLOYR framework allows rapid deployment of models, it assumes model training will occur using data sourced from Stanford's common data model. Additionally, DEPLOYR is currently configured to integrate specifically with EMR software provided by Epic Systems and would require additional API mappings and wrappers to transfer to other institutions using different EMR systems and clinical data warehouse structures.  Nevertheless, FHIR APIs were used when possible, which due to the 21st Century Cures Act have mandated support across U.S health institutions \cite{gordon202021st}. Further, the design decisions and considerations we have outlined in this study remain relevant to any institution attempting to install a deployment framework of their own.  DEPLOYR currently only supports deployment of ML applications trained using EMR data. While this can even include unstructured information represented in clinical notes, future upgrades would be necessary to support a broader array of clinical ML applications that use multiple data modalities \cite{rakha2021current, rezazade2021applications, eapen2020artificial}.

\section*{Conclusion}
There exist frameworks to govern and promote the implementation of accurate, actionable and reliable models that integrate with clinical workflow \cite{bedoya2022framework, reddy2020governance}. The creation of such frameworks has created the need for accompanying technical solutions that enable executing best practices laid out as desiderata \cite{blueprint}. To enable adherence to such guidance, we have developed and demonstrated DEPLOYR, a technical framework for rapidly deploying as well as prospectively monitoring custom, researcher developed ML models trained using clinical data warehouses. The framework's design elements and capabilities are demonstrated via silent deployment of multiple ML models triggered by button-clicks in a production instance of Epic, enabling prospective evaluation of the models' performance on unseen data directly in the deployment environment. The ability to perform randomized prospective evaluations of researcher created ML models supports the adoption of best practices that can close the ML implementation gap.


\paragraph{Acknowledgements and funding}
We used data provided by STARR, ``STAnford medicine Research data Repository'', a clinical data warehouse containing de-identified Epic data from Stanford Health Care (SHC), the University Healthcare Alliance (UHA) and Packard Children’s Health Alliance (PCHA) clinics and other auxiliary data from Hospital applications such as radiology PACS. The STARR platform is developed and operated by the Stanford Medicine Research IT team and is made possible by Stanford School of Medicine Research Office.  The content is solely the responsibility of the authors and does not necessarily represent the official views of the NIH or Stanford Health Care. Figures 1-5 were created with BioRender.com. This work was funded in part from the NIH/National Institute on Drug Abuse Clinical Trials Network (UG1DA015815–CTN-0136), Stanford Artificial Intelligence in Medicine and Imaging– Human-Centered Artificial Intelligence Partnership Grant, Doris Duke Charitable Foundation - Covid-19 Fund to Retain Clinical Scientists (20211260), Google Inc (in a research collaboration to leverage health data to predict clinical outcomes), and the American Heart Association - Strategically Focused Research Network - Diversity in Clinical Trials.

\paragraph{Conflict of interests}
JHC reported receiving consulting fees from Sutton Pierce and Younker Hyde MacFarlane PLLC and being a co-founder of Reaction Explorer LLC, a company that develops and licenses organic chemistry education software using rule-based artificial intelligence technology. CKC reported receiving consultation fees from Fountain Therapeutics. 

\paragraph{Code availability}
The DEPLOYR framework uses three distinct software repositories, \textit{DEPLOYR-dev} to train and validate models, \textit{DEPLOYR-serve} to expose models to the inference engine and \textit{DEPLOYR-dash} to monitor and visualize prospective model performance characteristics.  While this software is specifically designed to interact with Stanford data models and systems, we make \textit{DEPLOYR-dev}, \textit{DEPLOYR-dash} and an example of \textit{DEPLOYR-serve} publicly available for reference. 
\begin{itemize}
    \item 
    \href{https://github.com/HealthRex/deployr-dev}{\textit{DEPLOYR-dev}: https://github.com/HealthRex/deployr-dev}
    \item \href{https://github.com/HealthRex/deployr-serve}{\textit{DEPLOYR-serve}: https://github.com/HealthRex/deployr-serve}
    \item \href{https://github.com/HealthRex/deployr-dash}{\textit{DEPLOYR-dash}: https://github.com/HealthRex/deployr-dash}
\end{itemize}

\small

\bibliographystyle{vancouver}
\bibliography{main}

\newpage
\section*{Supplemental Note 1}
We created feature matrices for each cohort using count based representations \cite{rajkomar2018scalable}. For each cohort, a timeline of medical events was constructed from structured electronic medical record data available before inference time.  A mix of categorical and numerical data elements were considered. Categorical features included diagnosis (ICD 10) codes on a patient's problem list, medication orders and demographic variables including race and sex. Numerical features included prior laboratory results and the patient's age at inference time. Numerical features were discretized into tokens based on the percentile values they assumed in the training set distribution.  All numerical features were binned into five buckets. All diagnosis codes prior to prediction time were included in the patient's constructed timeline. Medication orders placed within 28 days of prediction time were included, as were laboratory results made available within 14 days of prediction time.  Sequences of tokens were then transformed into feature vectors in bag of words (counts) fashion.  The total number of features was 14145, 13631, and 14917 for the CBC, metabolic panel, and magnesium cohorts respectively. 
 Each cohort was split into training, validation and test sets based on the year the diagnostic order took place. Training sets included the years 2015 to 2019. Validation sets included orders taking place in 2020, and test sets included orders taking place in 2021. Random forest models were trained using the training sets, hyperparameters chosen using the validation sets, and performance evaluated on the test sets. We used the scikit-learn implementation of random forest models for these examples, though DEPLOYR allows any arbitrary model class or package.


\begin{figure}[b!]
\centering
\includegraphics[scale=0.18]{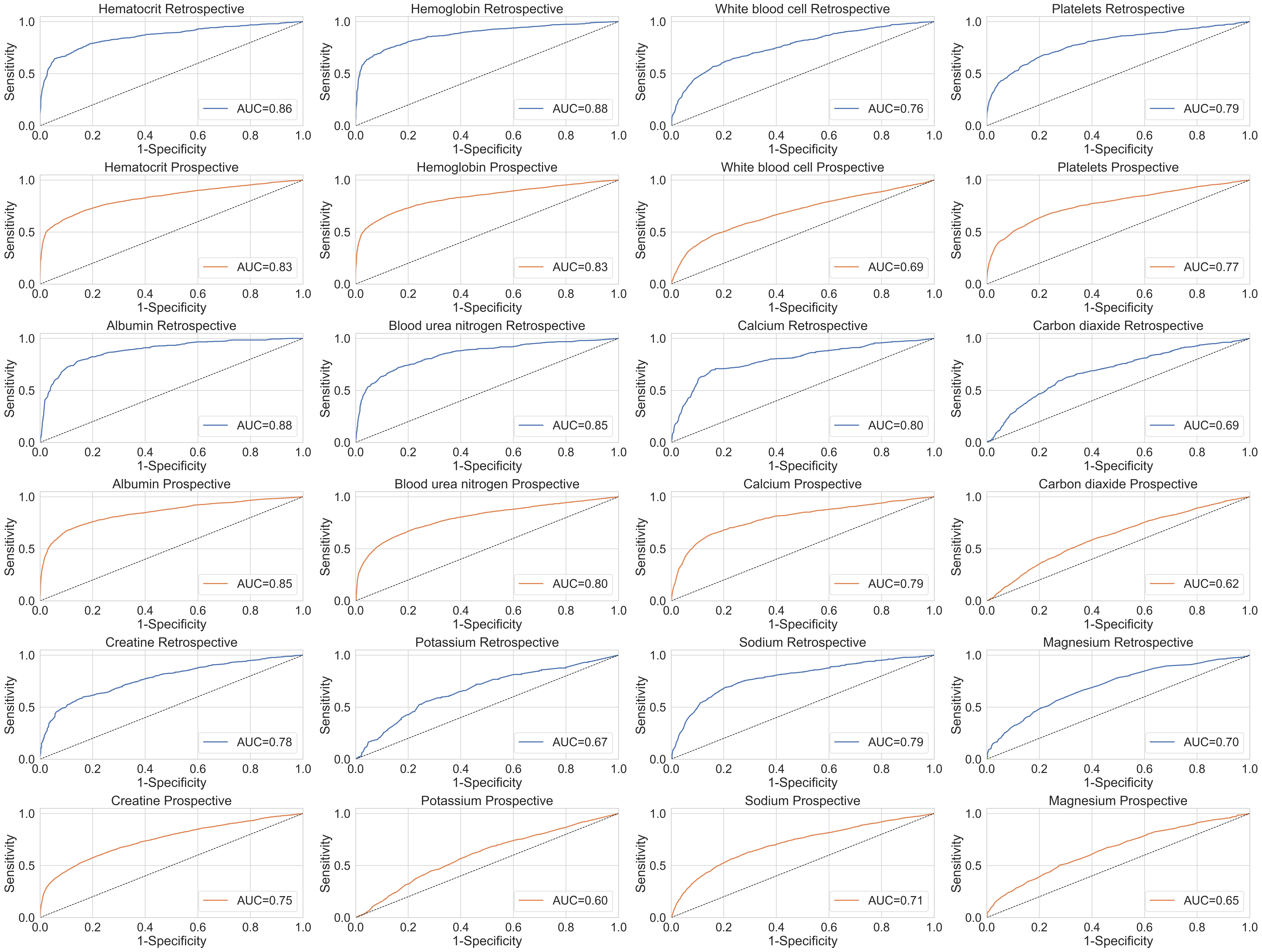}
\caption{Receiver operating characteristic curves for twelve deployed models on retrospective and prospective test sets.}
\label{fig1}
\end{figure}

\begin{figure}[t!]
\centering
\includegraphics[scale=0.18]{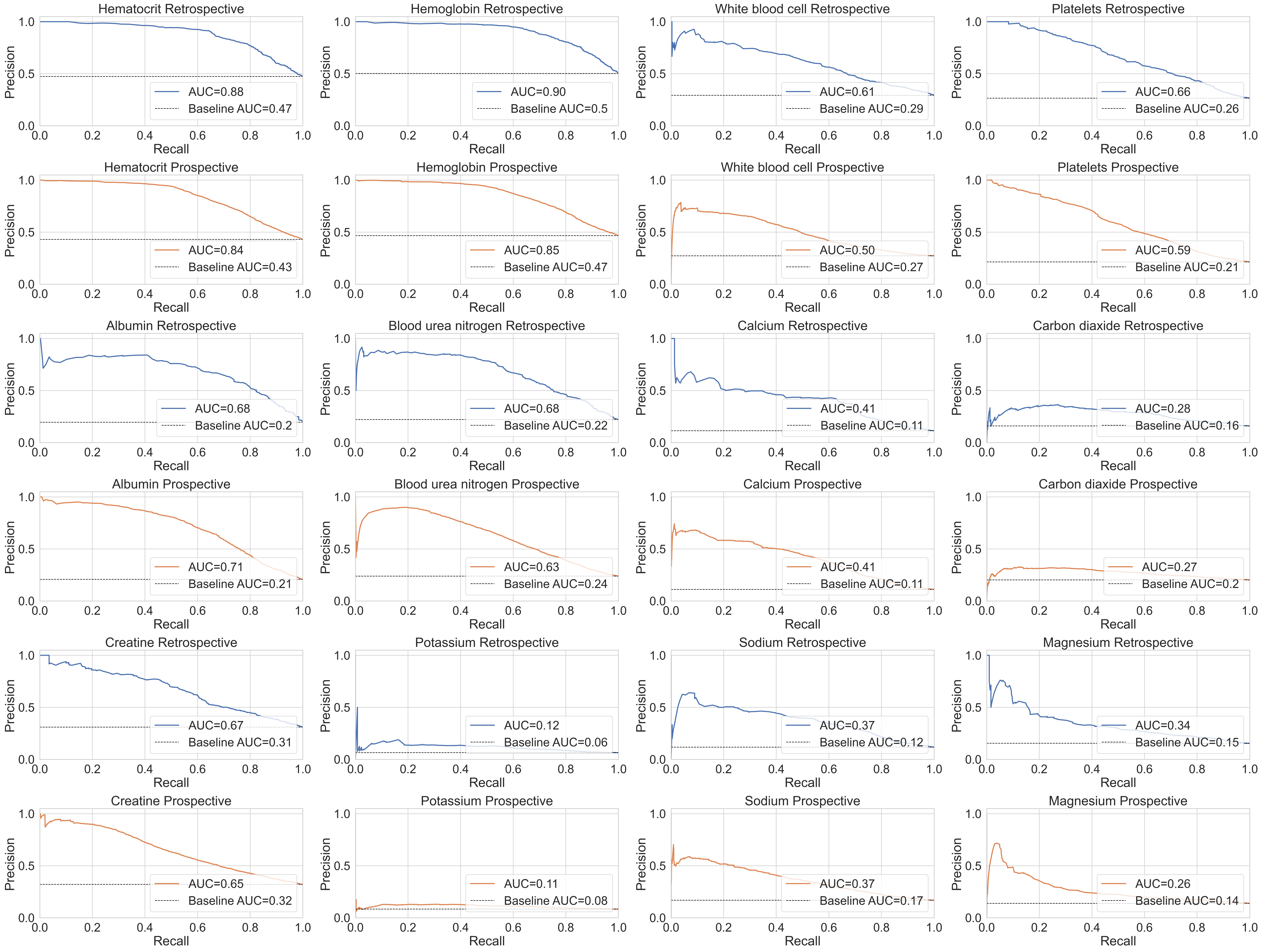}
\caption{Precision recall curves for twelve deployed models on retrospective and prospective test sets.}
\label{fig1}
\end{figure}

\begin{figure}[b!]
\centering
\includegraphics[scale=0.18]{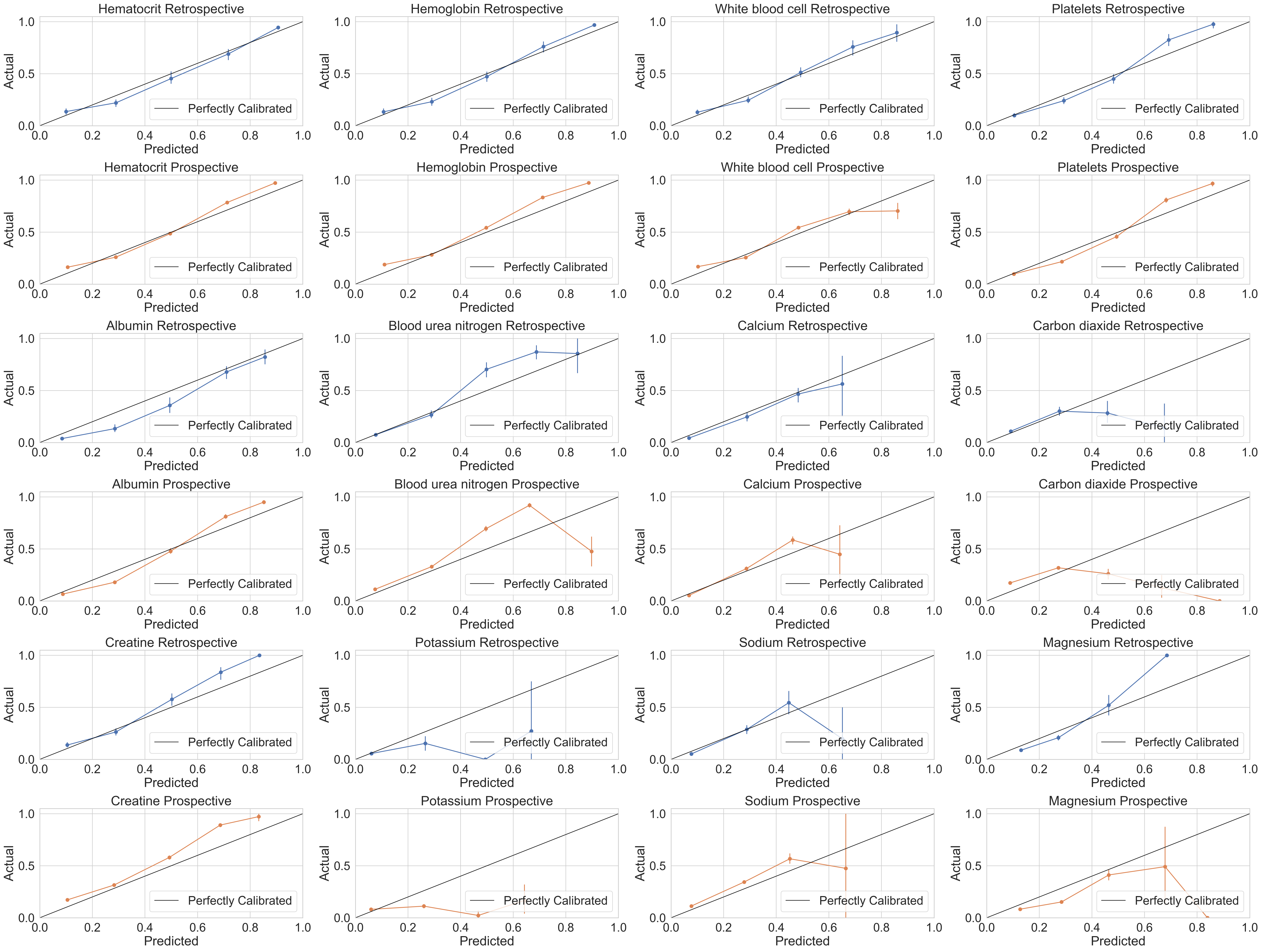}
\caption{Calibration plots for twelve deployed models on retrospective and prospective test sets.}
\label{fig1}
\end{figure}

\newpage

\begin{table}[b!]
\centering
\small
\caption{Hematocrit model performance by protected demographic groups}
\begin{tabular}{|l|l|l|l|}
\hline
\textbf{Prediction task}                       & \textbf{Group}                & \textbf{Retrospective AUROC}            & \textbf{Prospective AUROC}              \\ \hline
\multirow{12}{*}{\textbf{Hematocrit}}          & \textit{\textbf{Full cohort}} & \textit{\textbf{0.86 {[}0.85, 0.88{]}}} & \textit{\textbf{0.83 {[}0.83, 0.84{]}}} \\ \cline{2-4} 
                                               & sex\_Female                   & 0.85 {[}0.83, 0.88{]}                   & 0.83 {[}0.82, 0.84{]}                   \\ \cline{2-4} 
                                               & sex\_Male                     & 0.87 {[}0.84, 0.89{]}                   & 0.83 {[}0.83, 0.84{]}                   \\ \cline{2-4} 
                                               & race\_Asian                   & 0.85 {[}0.81, 0.88{]}                   & 0.85 {[}0.83, 0.86{]}                   \\ \cline{2-4} 
                                               & race\_Black                   & 0.84 {[}0.75, 0.91{]}                   & 0.79 {[}0.76, 0.81{]}                   \\ \cline{2-4} 
                                               & race\_Native American         & 0.87 {[}0.55, 1.00{]}                   & 0.81 {[}0.70, 0.90{]}                   \\ \cline{2-4} 
                                               & race\_Other                   & 0.85 {[}0.81, 0.89{]}                   & 0.82 {[}0.81, 0.83{]}                   \\ \cline{2-4} 
                                               & race\_Pacific Islander        & 0.91 {[}0.70, 1.00{]}                   & 0.85 {[}0.81, 0.89{]}                   \\ \cline{2-4} 
                                               & race\_Unknown                 & 0.80 {[}0.66, 0.90{]}                   & 0.75 {[}0.68, 0.81{]}                   \\ \cline{2-4} 
                                               & race\_White                   & 0.88 {[}0.85, 0.90{]}                   & 0.84 {[}0.83, 0.85{]}                   \\ \cline{2-4} 
                                               & age\_over\_40                 & 0.87 {[}0.85, 0.89{]}                   & 0.84 {[}0.83, 0.85{]}                   \\ \cline{2-4} 
                                               & sex\_Unknown                  & NaN                                     & 1.00 {[}1.00, 1.00{]}                   \\ \hline
\end{tabular}
\end{table}

\begin{table}[b!]
\centering
\small
\caption{Hemoglobin model performance by protected demographic groups}
\begin{tabular}{|l|l|l|l|}
\hline
\textbf{Prediction task}                       & \textbf{Group}                & \textbf{Retrospective AUROC}            & \textbf{Prospective AUROC}              \\ \hline
\multirow{12}{*}{\textbf{Hemoglobin}}          & \textit{\textbf{Full cohort}} & \textit{\textbf{0.88 {[}0.86, 0.89{]}}} & \textit{\textbf{0.83 {[}0.83, 0.84{]}}} \\ \cline{2-4} 
                                               & sex\_Female                   & 0.88 {[}0.86, 0.90{]}                   & 0.82 {[}0.81, 0.83{]}                   \\ \cline{2-4} 
                                               & sex\_Male                     & 0.88 {[}0.86, 0.90{]}                   & 0.84 {[}0.83, 0.85{]}                   \\ \cline{2-4} 
                                               & race\_Asian                   & 0.87 {[}0.84, 0.90{]}                   & 0.85 {[}0.84, 0.86{]}                   \\ \cline{2-4} 
                                               & race\_Black                   & 0.85 {[}0.77, 0.93{]}                   & 0.77 {[}0.74, 0.80{]}                   \\ \cline{2-4} 
                                               & race\_Native American         & 1.00 {[}1.00, 1.00{]}                   & 0.81 {[}0.71, 0.90{]}                   \\ \cline{2-4} 
                                               & race\_Other                   & 0.87 {[}0.84, 0.91{]}                   & 0.82 {[}0.80, 0.83{]}                   \\ \cline{2-4} 
                                               & race\_Pacific Islander        & 0.85 {[}0.59, 1.00{]}                   & 0.84 {[}0.80, 0.88{]}                   \\ \cline{2-4} 
                                               & race\_Unknown                 & 0.80 {[}0.67, 0.90{]}                   & 0.70 {[}0.63, 0.76{]}                   \\ \cline{2-4} 
                                               & race\_White                   & 0.89 {[}0.87, 0.91{]}                   & 0.84 {[}0.84, 0.85{]}                   \\ \cline{2-4} 
                                               & age\_over\_40                 & 0.89 {[}0.87, 0.91{]}                   & 0.84 {[}0.83, 0.85{]}                   \\ \cline{2-4} 
                                               & sex\_Unknown                  & NaN                                     & 1.00 {[}1.00, 1.00{]}                   \\ \hline
\end{tabular}
\end{table}

\begin{table}[b!]
\centering
\small
\caption{White blood cell model performance by protected demographic groups}
\begin{tabular}{|l|l|l|l|}
\hline
\textbf{Prediction task}                       & \textbf{Group}                & \textbf{Retrospective AUROC}            & \textbf{Prospective AUROC}              \\ \hline
\multirow{12}{*}{\textbf{White blood cell}}    & \textit{\textbf{Full cohort}} & \textit{\textbf{0.76 {[}0.74, 0.79{]}}} & \textit{\textbf{0.69 {[}0.68, 0.70{]}}} \\ \cline{2-4} 
                                               & sex\_Female                   & 0.77 {[}0.73, 0.80{]}                   & 0.68 {[}0.67, 0.69{]}                   \\ \cline{2-4} 
                                               & sex\_Male                     & 0.76 {[}0.73, 0.80{]}                   & 0.70 {[}0.69, 0.71{]}                   \\ \cline{2-4} 
                                               & race\_Asian                   & 0.76 {[}0.70, 0.82{]}                   & 0.70 {[}0.68, 0.72{]}                   \\ \cline{2-4} 
                                               & race\_Black                   & 0.75 {[}0.63, 0.86{]}                   & 0.68 {[}0.64, 0.72{]}                   \\ \cline{2-4} 
                                               & race\_Native American         & 0.50 {[}0.13, 0.88{]}                   & 0.68 {[}0.49, 0.82{]}                   \\ \cline{2-4} 
                                               & race\_Other                   & 0.73 {[}0.68, 0.78{]}                   & 0.67 {[}0.65, 0.69{]}                   \\ \cline{2-4} 
                                               & race\_Pacific Islander        & 1.00 {[}1.00, 1.00{]}                   & 0.76 {[}0.69, 0.83{]}                   \\ \cline{2-4} 
                                               & race\_Unknown                 & 0.87 {[}0.74, 0.96{]}                   & 0.69 {[}0.60, 0.76{]}                   \\ \cline{2-4} 
                                               & race\_White                   & 0.77 {[}0.73, 0.80{]}                   & 0.70 {[}0.68, 0.71{]}                   \\ \cline{2-4} 
                                               & age\_over\_40                 & 0.76 {[}0.73, 0.79{]}                   & 0.70 {[}0.69, 0.71{]}                   \\ \cline{2-4} 
                                               & sex\_Unknown                  & NaN                                     & 0.80 {[}0.40, 1.00{]}                   \\ \hline
\end{tabular}
\end{table}

\begin{table}[b!]
\centering
\small
\caption{Platelets model performance by protected demographic groups}
\begin{tabular}{|l|l|l|l|}
\hline
\textbf{Prediction task}                       & \textbf{Group}                & \textbf{Retrospective AUROC}            & \textbf{Prospective AUROC}              \\ \hline
\multirow{12}{*}{\textbf{Platelets}}           & \textit{\textbf{Full cohort}} & \textit{\textbf{0.79 {[}0.77, 0.82{]}}} & \textit{\textbf{0.77 {[}0.76, 0.78{]}}} \\ \cline{2-4} 
                                               & sex\_Female                   & 0.80 {[}0.77, 0.84{]}                   & 0.76 {[}0.75, 0.77{]}                   \\ \cline{2-4} 
                                               & sex\_Male                     & 0.78 {[}0.74, 0.81{]}                   & 0.78 {[}0.77, 0.79{]}                   \\ \cline{2-4} 
                                               & race\_Asian                   & 0.78 {[}0.72, 0.84{]}                   & 0.79 {[}0.77, 0.81{]}                   \\ \cline{2-4} 
                                               & race\_Black                   & 0.85 {[}0.75, 0.93{]}                   & 0.76 {[}0.72, 0.79{]}                   \\ \cline{2-4} 
                                               & race\_Native American         & 0.89 {[}0.67, 1.00{]}                   & 0.76 {[}0.60, 0.90{]}                   \\ \cline{2-4} 
                                               & race\_Other                   & 0.76 {[}0.70, 0.81{]}                   & 0.75 {[}0.73, 0.77{]}                   \\ \cline{2-4} 
                                               & race\_Pacific Islander        & 0.69 {[}0.07, 1.00{]}                   & 0.77 {[}0.69, 0.85{]}                   \\ \cline{2-4} 
                                               & race\_Unknown                 & 0.87 {[}0.72, 0.98{]}                   & 0.64 {[}0.51, 0.74{]}                   \\ \cline{2-4} 
                                               & race\_White                   & 0.80 {[}0.77, 0.83{]}                   & 0.78 {[}0.76, 0.79{]}                   \\ \cline{2-4} 
                                               & age\_over\_40                 & 0.78 {[}0.75, 0.81{]}                   & 0.77 {[}0.76, 0.78{]}                   \\ \cline{2-4} 
                                               & sex\_Unknown                  & NaN                                     & NaN                                     \\ \hline
\end{tabular}
\end{table}

\begin{table}[b!]
\centering
\small
\caption{Albumin model performance by protected demographic groups}
\begin{tabular}{|l|l|l|l|}
\hline
\textbf{Prediction task}                       & \textbf{Group}                & \textbf{Retrospective AUROC}            & \textbf{Prospective AUROC}              \\ \hline
\multirow{12}{*}{\textbf{Albumin}}             & \textit{\textbf{Full cohort}} & \textit{\textbf{0.88 {[}0.86, 0.91{]}}} & \textit{\textbf{0.85 {[}0.84, 0.86{]}}} \\ \cline{2-4} 
                                               & sex\_Female                   & 0.89 {[}0.86, 0.91{]}                   & 0.86 {[}0.85, 0.87{]}                   \\ \cline{2-4} 
                                               & sex\_Male                     & 0.88 {[}0.86, 0.91{]}                   & 0.84 {[}0.83, 0.85{]}                   \\ \cline{2-4} 
                                               & race\_Asian                   & 0.81 {[}0.74, 0.87{]}                   & 0.84 {[}0.82, 0.86{]}                   \\ \cline{2-4} 
                                               & race\_Black                   & 0.91 {[}0.80, 0.97{]}                   & 0.83 {[}0.78, 0.87{]}                   \\ \cline{2-4} 
                                               & race\_Native American         & 0.93 {[}0.61, 1.00{]}                   & 0.80 {[}0.69, 0.90{]}                   \\ \cline{2-4} 
                                               & race\_Other                   & 0.88 {[}0.83, 0.91{]}                   & 0.84 {[}0.82, 0.86{]}                   \\ \cline{2-4} 
                                               & race\_Pacific Islander        & 0.73 {[}0.46, 0.96{]}                   & 0.89 {[}0.83, 0.93{]}                   \\ \cline{2-4} 
                                               & race\_Unknown                 & 0.84 {[}0.63, 1.00{]}                   & 0.83 {[}0.75, 0.91{]}                   \\ \cline{2-4} 
                                               & race\_White                   & 0.91 {[}0.89, 0.93{]}                   & 0.86 {[}0.85, 0.87{]}                   \\ \cline{2-4} 
                                               & age\_over\_40                 & 0.89 {[}0.87, 0.91{]}                   & 0.87 {[}0.86, 0.88{]}                   \\ \cline{2-4} 
                                               & sex\_Unknown                  & NaN                                     & NaN                                     \\ \hline
\end{tabular}
\end{table}

\begin{table}[b!]
\centering
\small
\caption{Blood urea nitrogen model performance by protected demographic groups}
\begin{tabular}{|l|l|l|l|}
\hline
\textbf{Prediction task}                       & \textbf{Group}                & \textbf{Retrospective AUROC}            & \textbf{Prospective AUROC}              \\ \hline
\multirow{12}{*}{\textbf{Blood urea nitrogen}} & \textit{\textbf{Full cohort}} & \textit{\textbf{0.85 {[}0.83, 0.87{]}}} & \textit{\textbf{0.80 {[}0.79, 0.81{]}}} \\ \cline{2-4} 
                                               & sex\_Female                   & 0.84 {[}0.81, 0.88{]}                   & 0.79 {[}0.77, 0.80{]}                   \\ \cline{2-4} 
                                               & sex\_Male                     & 0.85 {[}0.82, 0.87{]}                   & 0.80 {[}0.78, 0.81{]}                   \\ \cline{2-4} 
                                               & race\_Asian                   & 0.86 {[}0.80, 0.92{]}                   & 0.81 {[}0.79, 0.83{]}                   \\ \cline{2-4} 
                                               & race\_Black                   & 0.85 {[}0.72, 0.95{]}                   & 0.82 {[}0.78, 0.85{]}                   \\ \cline{2-4} 
                                               & race\_Native American         & 1.00 {[}1.00, 1.00{]}                   & 0.91 {[}0.82, 0.97{]}                   \\ \cline{2-4} 
                                               & race\_Other                   & 0.85 {[}0.80, 0.88{]}                   & 0.78 {[}0.76, 0.80{]}                   \\ \cline{2-4} 
                                               & race\_Pacific Islander        & 0.91 {[}0.75, 1.00{]}                   & 0.83 {[}0.77, 0.88{]}                   \\ \cline{2-4} 
                                               & race\_Unknown                 & 0.85 {[}0.61, 0.99{]}                   & 0.74 {[}0.67, 0.82{]}                   \\ \cline{2-4} 
                                               & race\_White                   & 0.84 {[}0.81, 0.87{]}                   & 0.79 {[}0.78, 0.80{]}                   \\ \cline{2-4} 
                                               & age\_over\_40                 & 0.85 {[}0.82, 0.87{]}                   & 0.79 {[}0.78, 0.80{]}                   \\ \cline{2-4} 
                                               & sex\_Unknown                  & NaN                                     & NaN                                     \\ \hline
\end{tabular}
\end{table}

\begin{table}[b!]
\centering
\small
\caption{Calcium model performance by protected demographic groups}
\begin{tabular}{|l|l|l|l|}
\hline
\textbf{Prediction task}                       & \textbf{Group}                & \textbf{Retrospective AUROC}            & \textbf{Prospective AUROC}              \\ \hline
\multirow{12}{*}{\textbf{Calcium}}             & \textit{\textbf{Full cohort}} & \textit{\textbf{0.80 {[}0.76, 0.83{]}}} & \textit{\textbf{0.79 {[}0.78, 0.81{]}}} \\ \cline{2-4} 
                                               & sex\_Female                   & 0.76 {[}0.70, 0.81{]}                   & 0.78 {[}0.76, 0.80{]}                   \\ \cline{2-4} 
                                               & sex\_Male                     & 0.83 {[}0.79, 0.87{]}                   & 0.81 {[}0.79, 0.82{]}                   \\ \cline{2-4} 
                                               & race\_Asian                   & 0.76 {[}0.66, 0.86{]}                   & 0.78 {[}0.76, 0.81{]}                   \\ \cline{2-4} 
                                               & race\_Black                   & 0.80 {[}0.69, 0.91{]}                   & 0.75 {[}0.68, 0.81{]}                   \\ \cline{2-4} 
                                               & race\_Native American         & 0.78 {[}0.33, 1.00{]}                   & 0.81 {[}0.68, 0.93{]}                   \\ \cline{2-4} 
                                               & race\_Other                   & 0.84 {[}0.78, 0.89{]}                   & 0.82 {[}0.80, 0.84{]}                   \\ \cline{2-4} 
                                               & race\_Pacific Islander        & 0.87 {[}0.73, 1.00{]}                   & 0.81 {[}0.71, 0.89{]}                   \\ \cline{2-4} 
                                               & race\_Unknown                 & 0.64 {[}0.36, 0.89{]}                   & 0.78 {[}0.61, 0.93{]}                   \\ \cline{2-4} 
                                               & race\_White                   & 0.78 {[}0.73, 0.84{]}                   & 0.79 {[}0.77, 0.81{]}                   \\ \cline{2-4} 
                                               & age\_over\_40                 & 0.79 {[}0.75, 0.83{]}                   & 0.79 {[}0.78, 0.81{]}                   \\ \cline{2-4} 
                                               & sex\_Unknown                  & NaN                                     & NaN                                     \\ \hline
\end{tabular}
\end{table}

\begin{table}[b!]
\centering
\small
\caption{Carbon dioxide model performance by protected demographic groups}
\begin{tabular}{|l|l|l|l|}
\hline
\textbf{Prediction task}                       & \textbf{Group}                & \textbf{Retrospective AUROC}            & \textbf{Prospective AUROC}              \\ \hline
\multirow{12}{*}{\textbf{Carbon dioxide}}      & \textit{\textbf{Full cohort}} & \textit{\textbf{0.69 {[}0.66, 0.72{]}}} & \textit{\textbf{0.62 {[}0.61, 0.63{]}}} \\ \cline{2-4} 
                                               & sex\_Female                   & 0.67 {[}0.62, 0.71{]}                   & 0.61 {[}0.59, 0.62{]}                   \\ \cline{2-4} 
                                               & sex\_Male                     & 0.71 {[}0.66, 0.76{]}                   & 0.63 {[}0.61, 0.64{]}                   \\ \cline{2-4} 
                                               & race\_Asian                   & 0.73 {[}0.64, 0.80{]}                   & 0.61 {[}0.59, 0.64{]}                   \\ \cline{2-4} 
                                               & race\_Black                   & 0.76 {[}0.63, 0.88{]}                   & 0.62 {[}0.58, 0.66{]}                   \\ \cline{2-4} 
                                               & race\_Native American         & NaN                           & 0.74 {[}0.61, 0.84{]}                   \\ \cline{2-4} 
                                               & race\_Other                   & 0.72 {[}0.65, 0.77{]}                   & 0.63 {[}0.60, 0.65{]}                   \\ \cline{2-4} 
                                               & race\_Pacific Islander        & 0.26 {[}0.04, 0.54{]}                   & 0.69 {[}0.62, 0.76{]}                   \\ \cline{2-4} 
                                               & race\_Unknown                 & 0.96 {[}0.91, 1.00{]}                   & 0.64 {[}0.55, 0.73{]}                   \\ \cline{2-4} 
                                               & race\_White                   & 0.65 {[}0.60, 0.70{]}                   & 0.61 {[}0.59, 0.62{]}                   \\ \cline{2-4} 
                                               & age\_over\_40                 & 0.69 {[}0.65, 0.72{]}                   & 0.62 {[}0.61, 0.64{]}                   \\ \cline{2-4} 
                                               & sex\_Unknown                  & NaN                                     & 0.75 {[}0.25, 1.00{]}                   \\ \hline
\end{tabular}
\end{table}

\begin{table}[b!]
\centering
\small
\caption{Creatinine model performance by protected demographic groups}
\begin{tabular}{|l|l|l|l|}
\hline
\textbf{Prediction task}                       & \textbf{Group}                & \textbf{Retrospective AUROC}            & \textbf{Prospective AUROC}              \\ \hline
\multirow{12}{*}{\textbf{Creatinine}}            & \textit{\textbf{Full cohort}} & \textit{\textbf{0.78 {[}0.75, 0.80{]}}} & \textit{\textbf{0.75 {[}0.74, 0.76{]}}} \\ \cline{2-4} 
                                               & sex\_Female                   & 0.76 {[}0.73, 0.80{]}                   & 0.74 {[}0.73, 0.75{]}                   \\ \cline{2-4} 
                                               & sex\_Male                     & 0.77 {[}0.74, 0.81{]}                   & 0.75 {[}0.74, 0.76{]}                   \\ \cline{2-4} 
                                               & race\_Asian                   & 0.79 {[}0.74, 0.85{]}                   & 0.75 {[}0.73, 0.77{]}                   \\ \cline{2-4} 
                                               & race\_Black                   & 0.85 {[}0.77, 0.93{]}                   & 0.76 {[}0.73, 0.80{]}                   \\ \cline{2-4} 
                                               & race\_Native American         & 1.00 {[}1.00, 1.00{]}                   & 0.93 {[}0.87, 0.98{]}                   \\ \cline{2-4} 
                                               & race\_Other                   & 0.80 {[}0.75, 0.84{]}                   & 0.72 {[}0.71, 0.74{]}                   \\ \cline{2-4} 
                                               & race\_Pacific Islander        & 0.67 {[}0.43, 0.88{]}                   & 0.81 {[}0.75, 0.86{]}                   \\ \cline{2-4} 
                                               & race\_Unknown                 & 0.62 {[}0.43, 0.80{]}                   & 0.71 {[}0.64, 0.78{]}                   \\ \cline{2-4} 
                                               & race\_White                   & 0.75 {[}0.72, 0.79{]}                   & 0.75 {[}0.74, 0.76{]}                   \\ \cline{2-4} 
                                               & age\_over\_40                 & 0.78 {[}0.75, 0.80{]}                   & 0.76 {[}0.75, 0.77{]}                   \\ \cline{2-4} 
                                               & sex\_Unknown                  & NaN                                     & 0.88 {[}0.50, 1.00{]}                   \\ \hline
\end{tabular}
\end{table}

\begin{table}[b!]
\centering
\small
\caption{Potassium model performance by protected demographic groups}
\begin{tabular}{|l|l|l|l|}
\hline
\textbf{Prediction task}                       & \textbf{Group}                & \textbf{Retrospective AUROC}            & \textbf{Prospective AUROC}              \\ \hline
\multirow{12}{*}{\textbf{Potassium}}           & \textit{\textbf{Full cohort}} & \textit{\textbf{0.67 {[}0.61, 0.72{]}}} & \textit{\textbf{0.60 {[}0.59, 0.62{]}}} \\ \cline{2-4} 
                                               & sex\_Female                   & 0.65 {[}0.57, 0.72{]}                   & 0.60 {[}0.58, 0.62{]}                   \\ \cline{2-4} 
                                               & sex\_Male                     & 0.68 {[}0.62, 0.74{]}                   & 0.60 {[}0.58, 0.63{]}                   \\ \cline{2-4} 
                                               & race\_Asian                   & 0.75 {[}0.66, 0.83{]}                   & 0.61 {[}0.57, 0.65{]}                   \\ \cline{2-4} 
                                               & race\_Black                   & 0.79 {[}0.66, 0.90{]}                   & 0.56 {[}0.50, 0.62{]}                   \\ \cline{2-4} 
                                               & race\_Native American         & NaN                                     & 0.59 {[}0.46, 0.72{]}                   \\ \cline{2-4} 
                                               & race\_Other                   & 0.64 {[}0.55, 0.74{]}                   & 0.58 {[}0.55, 0.61{]}                   \\ \cline{2-4} 
                                               & race\_Pacific Islander        & 0.45 {[}0.06, 0.87{]}                   & 0.56 {[}0.47, 0.64{]}                   \\ \cline{2-4} 
                                               & race\_Unknown                 & 0.32 {[}0.10, 0.82{]}                   & 0.64 {[}0.46, 0.80{]}                   \\ \cline{2-4} 
                                               & race\_White                   & 0.65 {[}0.57, 0.72{]}                   & 0.61 {[}0.58, 0.63{]}                   \\ \cline{2-4} 
                                               & age\_over\_40                 & 0.65 {[}0.59, 0.71{]}                   & 0.60 {[}0.59, 0.62{]}                   \\ \cline{2-4} 
                                               & sex\_Unknown                  & NaN                                     & 0.00 {[}0.00, 0.00{]}                   \\ \hline
\end{tabular}
\end{table}

\begin{table}[b!]
\centering
\small
\caption{Sodium model performance by protected demographic groups}
\begin{tabular}{|l|l|l|l|}
\hline
\textbf{Prediction task}                       & \textbf{Group}                & \textbf{Retrospective AUROC}            & \textbf{Prospective AUROC}              \\ \hline
\multirow{12}{*}{\textbf{Sodium}}              & \textit{\textbf{Full cohort}} & \textit{\textbf{0.79 {[}0.75, 0.82{]}}} & \textit{\textbf{0.71 {[}0.70, 0.72{]}}} \\ \cline{2-4} 
                                               & sex\_Female                   & 0.78 {[}0.73, 0.83{]}                   & 0.71 {[}0.69, 0.72{]}                   \\ \cline{2-4} 
                                               & sex\_Male                     & 0.79 {[}0.74, 0.83{]}                   & 0.72 {[}0.70, 0.73{]}                   \\ \cline{2-4} 
                                               & race\_Asian                   & 0.81 {[}0.73, 0.87{]}                   & 0.73 {[}0.70, 0.75{]}                   \\ \cline{2-4} 
                                               & race\_Black                   & 0.78 {[}0.65, 0.89{]}                   & 0.71 {[}0.65, 0.75{]}                   \\ \cline{2-4} 
                                               & race\_Native American         & 0.73 {[}0.14, 1.00{]}                   & 0.74 {[}0.59, 0.89{]}                   \\ \cline{2-4} 
                                               & race\_Other                   & 0.78 {[}0.72, 0.84{]}                   & 0.68 {[}0.65, 0.70{]}                   \\ \cline{2-4} 
                                               & race\_Pacific Islander        & 0.91 {[}0.81, 1.00{]}                   & 0.70 {[}0.62, 0.78{]}                   \\ \cline{2-4} 
                                               & race\_Unknown                 & 0.72 {[}0.45, 0.99{]}                   & 0.75 {[}0.65, 0.84{]}                   \\ \cline{2-4} 
                                               & race\_White                   & 0.78 {[}0.72, 0.83{]}                   & 0.72 {[}0.71, 0.74{]}                   \\ \cline{2-4} 
                                               & age\_over\_40                 & 0.77 {[}0.73, 0.81{]}                   & 0.71 {[}0.70, 0.72{]}                   \\ \cline{2-4} 
                                               & sex\_Unknown                  & NaN                                     & 0.20 {[}0.00, 0.60{]}                   \\ \hline
\end{tabular}
\end{table}

\begin{table}[b!]
\centering
\small
\caption{Magnesium model performance by protected demographic groups}
\begin{tabular}{|l|l|l|l|}
\hline
\textbf{Prediction task}                       & \textbf{Group}                & \textbf{Retrospective AUROC}            & \textbf{Prospective AUROC}              \\ \hline
\multirow{12}{*}{\textbf{Magnesium}}           & \textit{\textbf{Full cohort}} & \textit{\textbf{0.70 {[}0.67, 0.73{]}}} & \textit{\textbf{0.65 {[}0.63, 0.67{]}}} \\ \cline{2-4} 
                                               & sex\_Female                   & 0.71 {[}0.66, 0.75{]}                   & 0.67 {[}0.63, 0.70{]}                   \\ \cline{2-4} 
                                               & sex\_Male                     & 0.70 {[}0.66, 0.74{]}                   & 0.64 {[}0.61, 0.67{]}                   \\ \cline{2-4} 
                                               & sex\_Unknown                  & NaN                                     & NaN                                     \\ \cline{2-4} 
                                               & race\_Asian                   & 0.76 {[}0.66, 0.84{]}                   & 0.66 {[}0.61, 0.72{]}                   \\ \cline{2-4} 
                                               & race\_Black                   & 0.69 {[}0.53, 0.82{]}                   & 0.70 {[}0.62, 0.77{]}                   \\ \cline{2-4} 
                                               & race\_Native American         & NaN                                     & 0.53 {[}0.29, 0.75{]}                   \\ \cline{2-4} 
                                               & race\_Other                   & 0.69 {[}0.62, 0.74{]}                   & 0.65 {[}0.61, 0.70{]}                   \\ \cline{2-4} 
                                               & race\_Pacific Islander        & 0.86 {[}0.67, 0.99{]}                   & 0.60 {[}0.47, 0.72{]}                   \\ \cline{2-4} 
                                               & race\_Unknown                 & 0.64 {[}0.28, 0.94{]}                   & 0.42 {[}0.18, 0.67{]}                   \\ \cline{2-4} 
                                               & race\_White                   & 0.70 {[}0.65, 0.74{]}                   & 0.65 {[}0.62, 0.68{]}                   \\ \cline{2-4} 
                                               & age\_over\_40                 & 0.70 {[}0.66, 0.74{]}                   & 0.65 {[}0.63, 0.68{]}                   \\ \hline
\end{tabular}
\end{table}

\end{document}